\documentclass{bmvc2k}

\title{Pyramid Attention Network \\for Semantic Segmentation}

\addauthor{Hanchao Li}{lihanchao@bit.edu.com}{1}
\addauthor{Pengfei Xiong}{xiongpengfei@megvii.com}{2}
\addauthor{Jie An}{jie.an@pku.edu.cn}{3}
\addauthor{Lingxue Wang*}{neobull@bit.edu.cn}{1}

\addinstitution{
	School of Optics and Photonics\\
	Beijing Institute of Technology\\
	Beijing, China \\
	(*Corresponding author)
}
\addinstitution{
	Megvii Inc. (Face++) \\
	Beijing, China
}
\addinstitution{
	School of Mathematical Sciences\\
	Peking University \\
	Beijing, China
}

\runninghead{Li ET AL.}{Pyramid Attention Network for Semantic Segmentation}

\begin{document}

\maketitle

\begin{abstract}
A \textit{Pyramid Attention Network}(PAN) is proposed to exploit the impact of global contextual information in semantic segmentation. Different from most existing works, we combine attention mechanism and spatial pyramid to extract precise dense features for pixel labeling instead of complicated dilated convolution and artificially designed decoder networks. Specifically, we introduce a Feature Pyramid Attention module to perform spatial pyramid attention structure on high-level output and combine global pooling to learn a better feature representation, and a Global Attention Upsample module on each decoder layer to provide global context as a guidance of low-level features to select category localization details. The proposed approach achieves state-of-the-art performance on PASCAL VOC 2012 and Cityscapes benchmarks with a new record of mIoU accuracy 84.0\% on PASCAL VOC 2012, while training without COCO dataset.
\end{abstract}

\section{Introduction}
\label{sec:intro}

\begin{figure*}
	\begin{center}
		\includegraphics[width=0.8\textwidth]{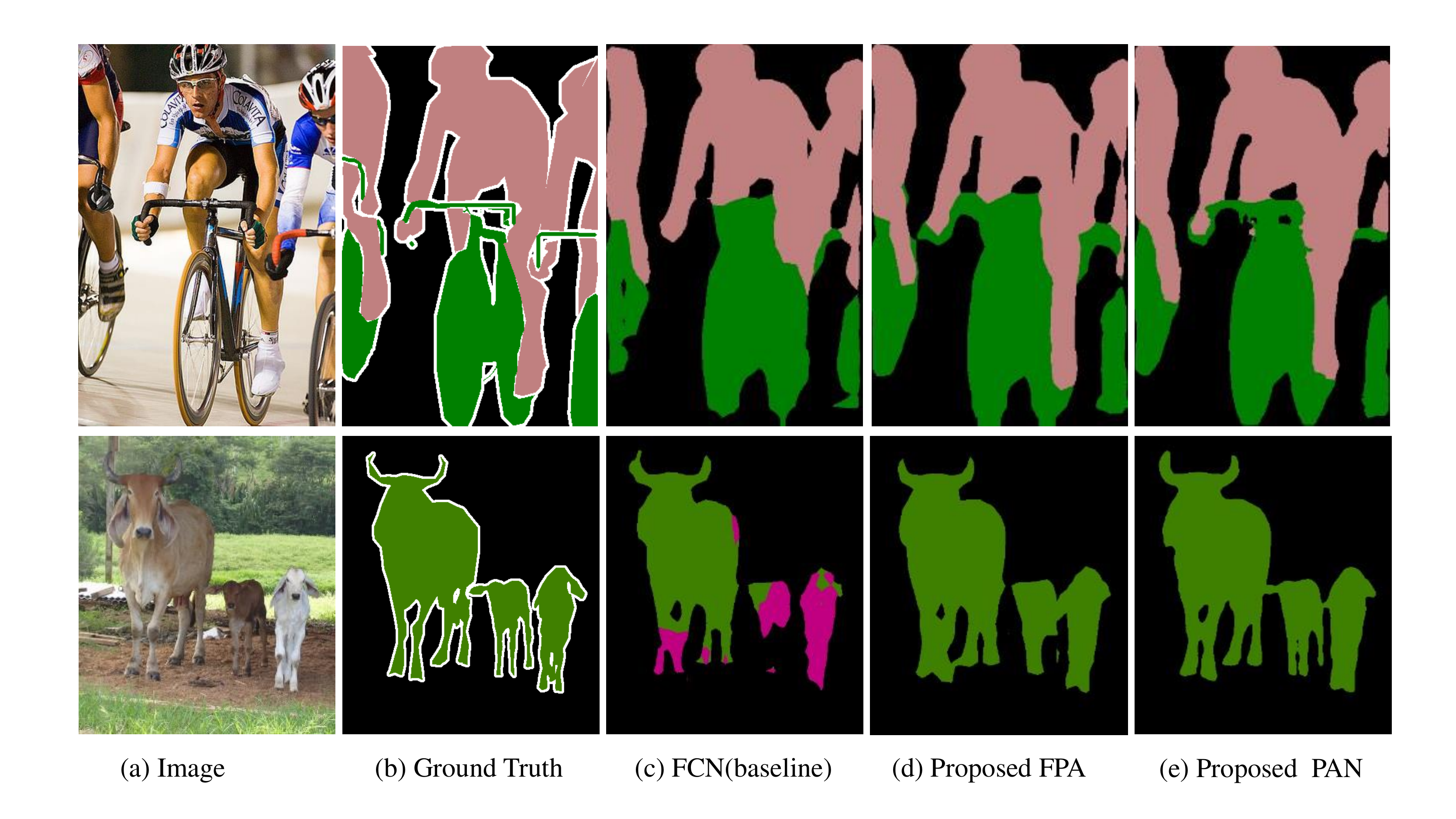}
	\end{center}
	\caption{Visualization results on VOC dataset\cite{everingham2010pascal}. 
		As we can see, FCN baseline model has difficulty in making predictions on small parts of objects and details. On the first row the bicycle handle is missing and the animal is predicted to be another wrong category on the second row.
		Our Feature Pyramid Attention(FPA) module and Global Attention Upsample(GAU) module are designed to increase receptive field and recover pixel localization details effectively.}
	\label{fig:img_show}
\end{figure*}

With the recent development of convolutional neural networks (CNNs)\cite{krizhevsky2012imagenet}\cite{he2016deep}, remarkable progress has occurred in pixel-wise semantic segmentation tasks due to rich hierarchical features and end-to-end trainable framework\cite{long2015fully}\cite{liu2015semantic}\cite{badrinarayanan2017segnet}\cite{zhao2017pyramid}. However, during encoding high dimension representations, original pixel-wise scene context suffers spatial resolution loss. 
As shown in Figure \ref{fig:img_show}, the FCN baseline lacks ability to make prediction
on small parts. The sheep beside the cow is made another wrong category on the second row. And on the first row the bicycle handle is missing. 
We take two main challenges into consideration.
 
The first issue is that the existence of objects at multiple scales cause difficulty in classification of categories. To solve this problem, PSPNet\cite{zhao2017pyramid} or DeepLab system\cite{chen2017rethinking} performs spatial pyramid pooling at different gird scales or dilate rates(called Atrous Spatial Pyramid Pooling, or ASPP). In ASPP module dilated convolution is a kind of sparse calculation which may cause grid artifacts \cite{wang2017understanding}. The pyramid pooling module proposed in PSPNet may lose pixel-level localization information. Inspired by SENet\cite{hu2017squeeze} and Parsenet\cite{liu2015semantic}, we attempt to extract precise pixel-level attention for high-level features extracted from CNNs. As Figure \ref{fig:img_show} shows, our proposed Feature Pyramid Attention (FPA) module is capable to increase receptive field and classify small objects effecitvely.

Another issue is that high-level features are skilled in making category classification, while weak in restructuring original resolution binary prediction. 
Some kind of U-shape networks, such as SegNet\cite{badrinarayanan2017segnet}, Refinenet\cite{lin2017refinenet}, Tiramisu proposed in \cite{jegou2017one} and Large Kernel Matters\cite{peng2017large}  perform complicate decoder module which use low-level information to help high-level features recover images detail. However, they are time consuming.
To solve this issue, we proposed a effective decoder module named Global Attention Upsample(GAU), which can extract global context of high-level features as guidance to weight low-level feature information without causing too much computation burden.

In summary, there are three main contributions in our paper. Firstly, we propose a Feature Pyramid Attention module to embed different scale context features in an FCN based pixel prediction framework. Then, We develop Global Attention Upsample, an effective decoder module for semantic segmentation. Lastly combining Feature Pyramid Attention module and Global Attention Upsample, our Pyramid Attention Network architecture archieves state-of-the-art accuracy on VOC2012 and cityscapes benchmark.

\section{Related Work}

In the following section, we review recent developments in semantic segmentation tasks. Since models based on Fully Convolutional Networks(FCNs)\cite{long2015fully} have achieved significant improvement on several segmentation tasks\cite{long2015fully}\cite{badrinarayanan2017segnet}. There is a lot of research focused on exploring the network structure to make better use of the contextual information\cite{zhao2017pyramid}\cite{zhang2018context}\cite{chen2017rethinking}. 

\noindent\textbf{Encoder-decoder:} State-of-the-art segmentation frameworks are mostly
based on encoder-decoder networks\cite{ronneberger2015u}\cite{badrinarayanan2017segnet}\cite{yu2018learning}\cite{peng2017large}, which have also been successfully applied to many computer vision tasks, including human pose estimation\cite{newell2016stacked}, object detection\cite{lin2017feature}\cite{liu2018path}, image stylization\cite{shrivastava2016beyond}, portrait matting\cite{shen2017automatic}\cite{shen2016automatic}, image Super-Resolution\cite{zhang2018residual}\cite{tong2017image}, and so-on. However, most methods attempt to combine the features of adjacent stages to enhance low-level features, without consideration of their diverse representation and the global context information.


\noindent\textbf{Global Context Attention:} Inspired from ParseNet\cite{liu2015parsenet}, global branch is adopted in several methods\cite{zhao2017pyramid}\cite{yu2018learning} to utilize global scene context. Global context easily enlarge the receptive field and enhance the consistency of pixel-wise classification. DFN\cite{yu2018learning} embeds global average pooling branch in the top to extend the U-shape architecture to a V-shape architecture. EncNet\cite{zhang2018context} introduces an encoding layer with a SENet\cite{hu2017squeeze} like module to capture the encoded semantics and predict scaling factors that are conditional on these encoded semantics. All of them results in great performance in different benchmarks. In this paper, we apply global pooling operator as an accessory module adding to the decoder branches to select the discriminative multi-resolution feature representations, which is proved to be effective.


\noindent\textbf{Spatial Pyramid:} This kind of models\cite{ghiasi2016laplacian}\cite{chen2017rethinking}\cite{zhao2017pyramid} apply in parallel spatial pyramid pooling to exploit the multi-scale context information. Spatial pyramid pooling \cite{he2014spatial}\cite{yu2015multi} has been widely employed to provide a good descriptor for overall scene interpretation, especially for various objects in multiple scales. Based on this, PSPNet\cite{zhao2017pyramid} and Deeplab series\cite{chen2018deeplab}\cite{chen2017rethinking} extend the global pooling module to the Spatial Pyramid Pooling and Atrous Spatial Pyramid Pooling, respectively. However, these models have shown high quality segmentation results on several benchmarks while usually need huge computing resources. 



\section{Method}
\label{sec:method}

In this section, we first introduce our proposed Feature Pyramid Attention(FPA) module and  Global Attention Upsample(GAU) module. Then we describe our complete encoder-decoder network architecture, Pyramid Attention Network designed for semantic segmentation task.

\begin{figure*}[htbp]
	\begin{center}
		\includegraphics[width=0.9\textwidth]{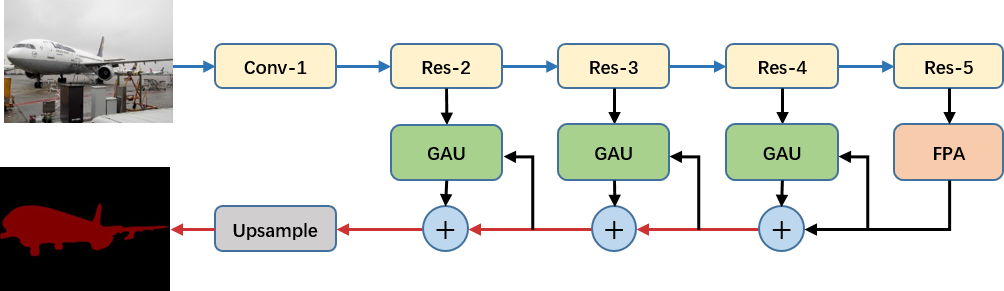}
	\end{center}
	\caption{Overview of the Pyramid Attention Network. We use ResNet-101 to extract dense features. Then we perform FPA and GAU to extract precise pixel prediction and localization details. The blue and red lines represent the downsample and upsample operators respectively.}
	\label{fig:wholenet}
\end{figure*}

\subsection{Feature Pyramid Attention}
\label{sec:FPA}


Recent models, such as PSPNet\cite{zhao2017pyramid} or DeepLab\cite{chen2017rethinking}, performed spatial pyramid pooling at several grid scales or apply ASPP module.
Dilated convolution 
may result in local information missing and `grids' which could be harmful for the local consistency of feature maps. 
Pyramid pooling module proposed in PSPNet loses pixel localization during different scale pooling operations.


\begin{figure*}[htbp]
	\begin{center}
		\includegraphics[width=1\textwidth]{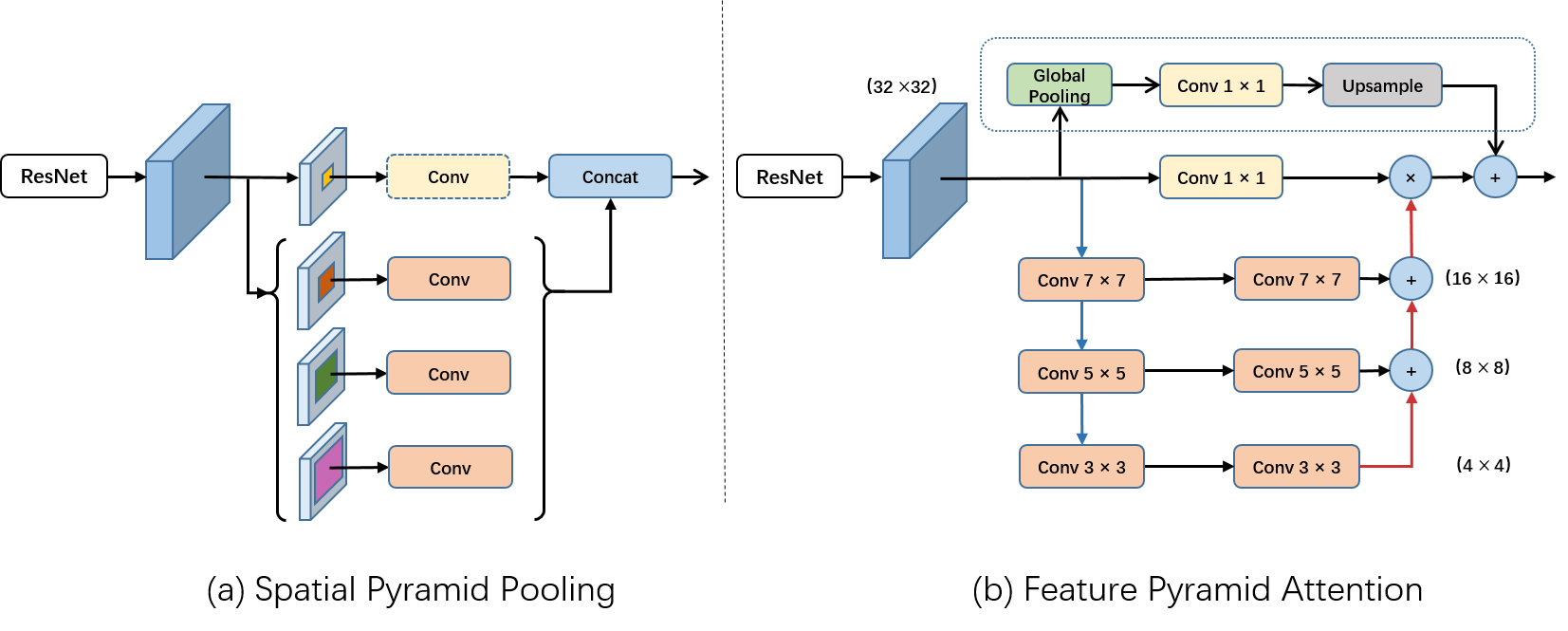}
	\end{center}
	\caption{Feature Pyramid Attention module structure. (a) Spatial Pyramid Pooling structure. (b) Feature Pyramid Attention module. '$4\times4$, $8\times8$, $16\times16$, $32\times32$' means the resolution of feature map. The dotted box means the global pooling branch. The blue and red lines represent the downsample and upsample operators respectively. Note that all Convolution layers are followed by batch normalization.}
	\label{fig:FPA}
\end{figure*}

Inspired by Attention Mechanism, we consider how to provide precise pixel-level attention for high-level features extracted from CNNs. In the current semantic segmentation architecture, the pyramid structure can extract different scale of feature information and increase receptive field effectively in pixel-level, while this kind of structure 
lacks global context prior attention to select the features channel-wise as in SENet\cite{hu2017squeeze} and EncNet\cite{zhang2018context}. On the other hand, using channel-wise attention vector is not enough to extract multi-scale features effectively and lack pixel-wise information.

With above observation, we propose Feature Pyramid Attention (FPA) module.
The pyramid attention module fuses features from under three different pyramid scales by implementing a U-shape structure like Feature Pyramid Network. To better extract context from different pyramid scales, we use $3\times3$, $5\times5$, $7\times7$ convolution in pyramid structure respectively. Since the resolution of high-level feature maps is small, using large kernel size doesn't bring too much computation burden. Then the pyramid structure integrates information of different scales step-by-step, which can incorporate neighbor scales of context features more precisely. Then the origin features from CNNs is multiplied pixel-wisely by the pyramid attention features after passing through a $1\times1$ convolution.
We also introduce global average pooling branch adding with the output features, which improve our FPA module performance further. The final module structure is shown in Figure \ref{fig:FPA}.

Benefiting from spatial pyramid structure, Feature Pyramid Attention module can fuse different scale context information and produce better pixel-level attention for high-level feature maps in the meantime. Unlike PSPNet or ASPP concatenates different pyramid scale feature maps before channel reduce convolution layer, our context information is multiplied with original feature map pixel-wisely, which doesn't introduce too much computation.

\subsection{Global Attention Upsample}

There are several decoder architecture designs in the current semantic segmentation networks. PSPNet\cite{zhao2017pyramid} or Deeplab\cite{chen2017rethinking} uses bilinearly upsample directly which can be seen as a naive decoder. DUC\cite{wang2017understanding} uses large channel convolution combined with reshaping as one-step decoder module. Both the naive decoder and one-step decoder lack different scales of low-level feature map information and could be harmful to recover spatial localization to origin resolution.
The common encoder-decoder networks mainly consider using different scales of feature information and gradually recover sharp object boundaries in the decoder path. In addition, most methods of this type usually use complicate decoder blocks, which cost plenty of computation resource. 

Recent research  has shown that combining CNNs with well-designed pyramid module can obtain considerable performance and capability to obtain category information. We consider that the main character of decoder module is to repair category pixel localization. Furthermore, high-level features with abundant category information can be used to weight low-level information to select precise resolution details.

\begin{figure*}[htbp]
	\begin{center}
		\includegraphics[width=0.6\textwidth]{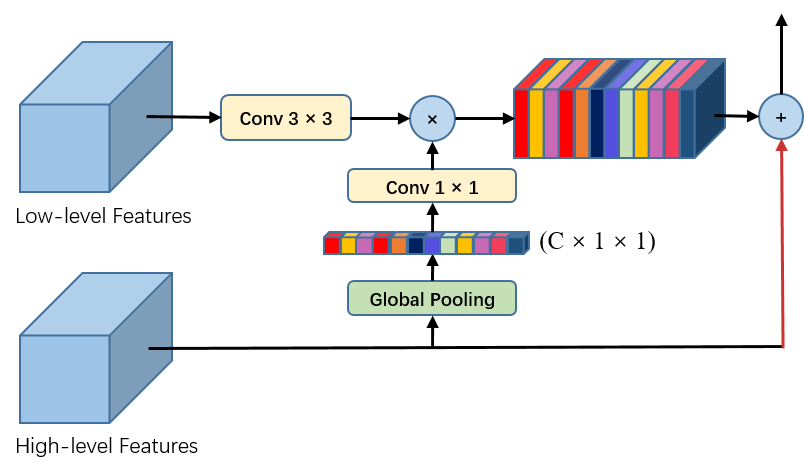}
	\end{center}
	\caption{Global Attention Upsample module structure}
	\label{fig:GAU}
\end{figure*}

Our Global Attention Upsample module performs global average pooling to provide global context as a guidance of low-level features to select category localization details. In detail, we perform $3\times3$ convolution on the low-level features to reduce channels of feature maps from CNNs. The global context generated from high-level features is through a $1\times1$ convolution with batch normalization and ReLU non-linearity, then multiplied by the low-level features. Finally high-level features are added with the weighted low-level features and upsampled gradually. This module deploys different scale feature maps more effectively and uses high-level features provide guidance information to low-level feature maps in a simple way.

\subsection{Network Architecture}

With proposed Feature Pyramid Attention(FPA) and Global Attention Upsample(GAU), we propose our Pyramid Attention Network(PAN) as Figure \ref{fig:wholenet}. We use ResNet-101 pretrained on ImageNet with the dilated convolution strategy to extract the feature map. In detail, the dilated convolution with rate of 2 is applied to \textit{res5b} blocks, so the output size of feature maps from ResNet is 1/16 of the input image like DeepLabv3+. We also replace the 7x7 convolutional layer in the original ResNet-101 by three $3\times3$ convolutional layers like PSPNet and DUC. We use the FPA module to gather dense pixel-level attention information from the output of ResNet. Combined with global context, the final logits are follow by GAU module to generate the final prediction maps.

We treat Feature Pyramid Attention module as center block between Encoder and Decoder structure. Without Global Attention Upsample module, Feature Pyramid Attention module can also provide enough precise pixel-level prediction and class identification, as we show in Section \ref{sec:exper} below. After implementing Feature Pyramid Attention module, we perform Global Attention Module as a fast and effective decoder structure, which use high-level features to guide low-level information and combine both precisely.

\section{Experimental Results}
\label{sec:exper}

We evaluate our approach on two main segmentation datasets: PASCAL VOC 2012 semantic segmentation\cite{everingham2010pascal} and urban scene dataset Cityscapes\cite{cordts2016cityscapes}. 
We first conduct a complete ablation study on VOC 2012 dataset, and finally report the state-of-art performances. 

For a practical deep learning system, devil is always in
the details. We use the "poly" learning rate policy where the initial rate is multiplied by $(1-\frac{iter}{max\_iter})^{power}$ with power 0.9 and initial rate $4e-−3$, and train the network using mini-batch stochastic gradient descent (SGD) with batch size 16, momentum 0.9 and weight decay 0.0001. The cross-entropy error at each pixel over the categories is applied as our loss function. We adopt randomly left-right flipping and random scaling between 0.5 and 2 for all datasets during training.

\begin{table}
	\begin{center}
		\begin{tabular}{|l|c|c|}
			\hline
			Method & mean IoU(\%) & Pixel Acc.(\%) \\
			\hline\hline
			ResNet101        & 72.60 & 93.90 \\
			ResNet101+SE              & 75.74 & 94.48 \\
			ResNet101+C333+MAX        & 77.13 & 94.79 \\
			ResNet101+C333+AVE        & 77.54 & 94.89 \\
			ResNet101+C333+MAX+GP     & 77.29 & 94.81 \\
			ResNet101+C333+AVE+GP     & 77.61 & 94.88 \\
			\hline
			ResNet101+C357+MAX        & 77.40 & 94.77 \\
			ResNet101+C357+AVE        & 78.19 & 95.00 \\
			ResNet101+C357+MAX+GP     & 77.03 & 94.71 \\
			ResNet101+C357+AVE+GP     & 78.37 & 95.03 \\
			\hline
		\end{tabular}
	\end{center}
	\caption{Detailed performance of Feature Pyramid Attention with different settings.
	`SE' means using SENet attention module to replace the pyramid structure. For the pyramid structure used in Feature Pyramid Attention module, `C333' represent all the kernel size of convolution is {$3\times3$}. `C357' means the kernel size of convolution is {$3\times3$, $5\times5$, $7\times7$} respectively, as Figure \ref{fig:FPA} shown. `MAX' and `AVE' represent max pooling and average pooling operations. `GP' means the global pooling branch.}
	\label{fig:fpa-expr1}
\end{table}

\subsection{Ablation Experiments}
\label{sec:voc}

The PASCAL VOC 2012 contains 20 foreground object classes and one background class. The original dataset involves 1,464 images for training, 1,449 images for validation and 1,456 images for testing. The dataset is augmented by the Semantic Boundaries Dataset\cite{hariharan2015hypercolumns}, resulting in 10,582 images for training.  In this subsection, we use PASCAL VOC 2012 validation set for the evaluation and our crop size is $512\times512$. The performance is measured in terms of pixel intersection-over-union (IOU) averaged across the 21 classes with the single-scale input.

\subsubsection{Feature Pyramid Attention}

First, we experiment the performance of the base ResNet-101 with dilated convolution mentioned above and directly upsample the feature network's output. To evaluate our Feature Pyramid Attention(FPA) module, we perform FPA module after ResNet with direct upsampling as the same as the baseline. In detail, we conduct experiments with several setting, including pooling types of max and average, attention with pyramid structure or just one branch using global context similar to SENet, different kernel size in pyramid structure, with or without global pooling branch.

\noindent\textbf{Ablation for pooling type:} We notice that average pooling works better than max pooling in all settings. For using all the convolution with $3\times3$ kernel size, `AVE' setting improves the performance from 77.13\% to 77.54\% compared to `MAX' setting. So we adopt average pooling in our final module.

\noindent\textbf{Ablation for pyramid structure:} As shown in Table \ref{fig:fpa-expr1}, our baseline model achieves mIoU of 72.60\% on the validation set. Based on the observation in Section \ref{sec:method}, we firstly implement the pyramid structure with `C333' setting and average pooling , which improves the performance from 72.6\% to 77.54\%. We also perform the SENet attention module to replace the pyramid structure to evaluate the performance compared with our pyramid attention module. As shown in Table \ref{fig:fpa-expr1}, compared to SENet attention module, `C333' and `AVE' setting improves the performance by almost 1.8\%. 

\noindent\textbf{Ablation for kernel size:} For the pyramid structure using average pooling, we use large kernel convolution `C357' to replace $3\times3$ kernel size, shown in Figure \ref{fig:FPA}. As mentioned in Section \ref{sec:FPA}, the resolutions of feature maps in the pyramid structure are $16\times16$, $8\times8$, $4\times4$ respectively, so using large kernel setting doesn't bring too much calculation burden. This improves performance from 77.54\% to 78.19\%. 

\noindent\textbf{Ablation for global pooling:} We further add global pooling branch in the pyramid structure to improve performance. Finally, the best setting yields results 78.37/95.03 in terms of Mean IoU and Pixel Acc. (\%). The results show that our pyramid attention can extract pixel-level attention information effectively. 

When using directly upsampling as naive decoder, our FPA module also shows notable progress compared to PSPNet and DeepLabv3 under the same output stride (\textit{stride}=16). Since PSPNet didn't report performance on VOC validation set, we implement the pyramid pooling module proposed in PSPNet combined with our base model. We use the same training protocol and keep the same output stride for fair comparison.
The results are shown in Table \ref{fig:fpa-expr2}. Our FPA module is more capable than the other two modules. Compared to the pyramid pooling module and DeepLabv3, our FPA module outperforms DeepLabv3 by almost 1.2\% under the same output stride.

\begin{table}
	\begin{center}
		\begin{tabular}{|l|c|c|}
			\hline
			Method & mean IoU(\%) & Pixel Acc.(\%) \\
			\hline\hline
			ResNet101+PSPNet* \cite{zhao2017pyramid}                     & 76.82 & 94.62 \\  
			DeepLab V3\cite{chen2017rethinking} (\textit{stride}=16)   & 77.21 &   --  \\ 
			ResNet101+FPA                       & 78.37 & 95.03 \\
			\hline
		\end{tabular}
	\end{center}
	\caption{Comparison with other state-of-art methods. `ResNet101+PSPNet*' means that we implement the pyramid pooling module on our base model(\textit{stride}=16).}
	\label{fig:fpa-expr2}
\end{table}

\subsubsection{Global Attention Upsample}

Since Feature Pyramid Attention module provides precise pixel-level prediction, Global Attention Upsampling (GAU) focus on using low-level features to recover pixel localization. To be specific, we perform global pooling and $1\times1$ convolution used to generate global context as guidance. The $3\times3$ convolution used to reduce the channels of the low-level feature map from encoder module. 

We first evaluate GAU module combined with ResNet101 baseline, then we experiment FPA and GAU together on VOC 2012 \textit{val} set. As for the design of decoder module, we evaluate three different design respectively, (1) only using low-level features from skip-connection without global context attention branch, (2) using $1\times1$ convolution to reduce channels of low-level features in GAU module, (3) replace $1\times1$ convolution with $3\times3$ convolution to preform channel reduction. As shown in Table \ref{fig:GAU-expr1}, without global context attention branch, our decoder module merely improves performance from 72.60\% to 73.56\%. Then we add global pooling operation to extract global context attention information, which improves performance from 73.56\% to 77.84\% significantly.

\begin{table}[htbp]
	\begin{center}
		\begin{tabular}{|l|c|c|c|c|c|}
			\hline
			Method & GP  & $1\times1$ Conv & $3\times3$ Conv & mean IoU(\%) & Pixel Acc.(\%) \\
			\hline\hline
			ResNet101        &  --- &  --- & --- & 72.60 & 93.90 \\
			\hline
			ResNet101+GAU  &          &          & $\surd$ & 73.56 & 94.15 \\
			ResNet101+GAU  &  $\surd$ &  $\surd$ &         & 77.48 & 94.89 \\
			ResNet101+GAU  &  $\surd$ &          & $\surd$ & 77.84 & 94.96 \\
			\hline
		\end{tabular}
	\end{center}
	\caption{Detailed performance with different settings of decoder module.}
	\label{fig:GAU-expr1}
\end{table}

As we take GAU as a slight and effective decoder module, we also compare with Global Convolution Network\cite{peng2017large} and Discriminate Feature Network(DFN)\cite{yu2018learning}. The results are shown in Table \ref{fig:GAU-expr2}. The structure of `DFN(ResNet101+RRB)' is a ResNet101 combined with proposed Refinement Residual Block(RRB) as decoder module. Our decoder module outperforms RRB by 1.2\%. It is worth noted that Global Convolution Network used extra COCO dataset combined with VOC dataset for training and obtained 77.50\%, while our decoder module can achieve 77.84\% without COCO dataset for training.

\begin{table}[htbp]
	\begin{center}
		\begin{tabular}{|l|c|c|c|c|c|}
			\hline
			Method & mean IoU(\%) & Pixel Acc.(\%) \\
			\hline\hline
			DFN(Res-101+RRB) \cite{yu2018learning}      & 76.65 & --  \\
			Global Convolution Network \cite{peng2017large} & 77.50 & -- \\
			ResNet101+GAU   & 77.84 & 94.96 \\
			\hline
		\end{tabular}
	\end{center}
	\caption{Comparison with other state-of-art decoder module.}
	\label{fig:GAU-expr2}
\end{table}

\subsection{PASCAL VOC 2012}

Combined our best setting on Feature Pyramid Attention and Global Attention Upsampling module, we experiment the complete network architecture---Pyramid Attention Network (PAN) on PASCAL VOC 2012 \textit{test} set. In evaluation, we apply the multi-scale inputs (with scales$=\{0.5, 0.75, 1.0, 1.25, 1.5, 1.75\}$) and also left-right flipped the images in evaluation, as listed in Table \ref{fig:PAN-val}. Since the PASCAL VOC 2012 dataset provides higher quality of annotation than the augmented datasets [9], we further fine-tune our model on PASCAL VOC 2012 \textit{trainval} set for evaluation on the test set. More performance details are listed in Table \ref{fig:PAN-test}. Finally our proposed approach achieve performance of 84.0\% without MS-COCO \cite{lin2014microsoft} and Dense-CRF post-processing\cite{jampani2016learning}.

\begin{table}[htbp]
	\begin{center}
		\begin{tabular}{|l|c|c|c|c|}
			\hline
			Method  & MS & Flip & mean IoU(\%) & Pixel Acc.(\%) \\
			\hline\hline
			PAN &          &         & 79.38 & 95.25 \\
			PAN &  $\surd$ &         & 80.77 & 95.65 \\
			PAN &  $\surd$ & $\surd$ & 81.19 & 95.75 \\
			\hline
		\end{tabular}
	\end{center}
	\caption{Performance on VOC 2012 \textit{val} set.}
	\label{fig:PAN-val}
\end{table}

\begin{table}[htbp]
	\begin{center}
		\resizebox{\textwidth}{15mm}{
			\setlength{\tabcolsep}{0.8mm}{
		\begin{tabular}{l|cccccccccccccccccccc|c}
			\hline
			Method	& aero & bike & bird & boat & bottle & bus & car & cat & chair & cow & table & dog & horse & mbike & person & plant & sheep & sofa & train & tv & mean IoU(\%) \\
			\hline\hline
			FCN\cite{long2015fully}         & 76.8 & 34.2 & 68.9 & 49.4 & 60.3 & 75.3 & 74.7 & 77.6 & 21.4 & 62.5 & 46.8 & 71.8 & 63.9 & 76.5 & 73.9 & 45.2 & 72.4 & 37.4 & 70.9 & 55.1 & 62.2    \\
			DeepLabv2\cite{chen2018deeplab}   & 84.4 & 54.5 & 81.5 & 63.6 & 65.9 & 85.1 & 79.1 & 83.4 & 30.7 & 74.1 & 59.8 & 79.0 & 76.1 & 83.2 & 80.8 & 59.7 & 82.2 & 50.4 & 73.1 & 63.7 & 71.6   \\
			CRF-RNN\cite{zheng2015conditional}     & 87.5 & 39.0 & 79.7 & 64.2 & 68.3 & 87.6 & 80.8 & 84.4 & 30.4 & 78.2 & 60.4 & 80.5 & 77.8 & 83.1 & 80.6 & 59.5 & 82.8 & 47.8 & 78.3 & 67.1 & 72.0    \\
			DeconvNet\cite{noh2015learning}   & 89.9 & 39.3 & 79.7 & 63.9 & 68.2 & 87.4 & 81.2 & 86.1 & 28.5 & 77.0 & 62.0 & 79.0 & 80.3 & 83.6 & 80.2 & 58.8 & 83.4 & 54.3 & 80.7 & 65.0 & 72.5 \\
			DPN\cite{liu2015semantic}         & 87.7 & 59.4 & 78.4 & 64.9 & 70.3 & 89.3 & 83.5 & 86.1 & 31.7 & 79.9 & 62.6 & 81.9 & 80.0 & 83.5 & 82.3 & 60.5 & 83.2 & 53.4 & 77.9 & 65.0 & 74.1 \\
			Piecewise\cite{lin2016efficient}   & 90.6 & 37.6 & 80.0 & 67.8 & 74.4 & 92.0 & 85.2 & 86.2 & 39.1 & 81.2 & 58.9 & 83.8 & 83.9 & 84.3 & 84.8 & 62.1 & 83.2 & 58.2 & 80.8 & 72.3 & 75.3 \\
			PSPNet\cite{zhao2017pyramid}      & 91.8 & 71.9 & 94.7 & 71.2 & 75.8 & 95.2 & 89.9 & \textbf{95.9} & 39.3 & 90.7 & 71.7 & \textbf{90.5} & \textbf{94.5} & 88.8 & \textbf{89.6} & 72.8 & 89.6 & \textbf{64.0} & 85.1 & 76.3 & 82.6 \\
			EncNet\cite{zhang2018context}      & 94.1 & 69.2 & \textbf{96.3} & \textbf{76.7} & \textbf{86.2} & 96.3 & 90.7 & 94.2 & 38.8 & 90.7 & \textbf{73.3} & 90.0 & 92.5 & 88.8 & 87.9 & 68.7 & 92.6 & 59.0 & 86.4 & 73.4 & 82.9 \\
			\hline
			PAN(ours) 	            & \textbf{95.7} & \textbf{75.2} & 94.0 & 73.8 & 79.6 & \textbf{96.5} & \textbf{93.7} & 94.1 & \textbf{40.5} & \textbf{93.3} & 72.4 & 89.1 & 94.1 & \textbf{91.6} & 89.5 & \textbf{73.6} & \textbf{93.2} & 62.8 & \textbf{87.3} & \textbf{78.6} & 84.0 \\
			\hline
		\end{tabular}}
	}
	\end{center}
	\caption{Per-class results on PASCAL VOC 2012 \textit{test} set. PAN outperforms the state-of-art approaches and achieves 84.0\% without pre-training on COCO dataset.}
	\label{fig:PAN-test}
\end{table}

\subsection{Cityscapes}

The Cityscapes contains 30 classes, and 19 of them are considered for training and evaluation. The dataset contains 5,000 finely annotated images and 19,998 images with coarse annotation. 
In detail, the fine annotated images are split into training, validation and testing sets with 2,979, 500 and 1,525 images respectively. During training, we didn't use coarse annotation dataset. Our crop size of image is $768 \times 768$. We also used ResNet101 as base model like in Section \ref{sec:voc}. The performance on the test set are reported in Table \ref{fig:city-test}.

\begin{table}[htbp]
	\begin{center}
		\begin{tabular}{|l|c|}
			\hline
			Method      & mean IoU(\%) \\
			\hline\hline
			DPN \cite{liu2015semantic}        & 66.8  \\
			DeepLabv2 \cite{chen2018deeplab}     & 70.4  \\
			Piecewise \cite{lin2016efficient}  & 71.6  \\
			RefineNet \cite{lin2017refinenet}   & 73.6  \\
			DUC \cite{wang2017understanding}         & 77.6  \\
			PSPNet \cite{zhao2017pyramid}      & 78.4  \\
			\hline
			PAN(ours)        & 78.6  \\
			\hline
			\end{tabular}
	\end{center}
	\caption{Performance on Cityscapes testing set without coarse annotation dataset.}
	\label{fig:city-test}
\end{table}

\section{Conclusion}

We have proposed a notable Pyramid Attention Network for semantic segmentation. We designed Feature Pyramid Attention module and an effective decoder module Global Attention Upsample. Feature Pyramid Attention module provides pixel-level attention information and increases receptive field by performing pyramid structure. Global Attention Upsample module exploits high-level feature map to guide low-level features recovering pixel localization. Our experimental results show that the proposed approach can achieve comparable performance with other state-of-art models on the PASCAL VOC 2012 semantic image segmentation benchmark.

\section*{Acknowledgements}

This work is supported by the National Natural Science Foundation of China (Nos. 61471044).

\bibliography{egbib}

\end{document}